\documentclass[runningheads]{llncs}
\usepackage{graphicx}
\usepackage{amsmath,amssymb,amsfonts}
\usepackage{algorithmic}
\usepackage[caption=false]{subfig}
\usepackage{booktabs}
\begin{document}
\title{Sales Demand Forecast in E-commerce using a Long Short-Term Memory Neural Network Methodology}
%
\author{Kasun Bandara\inst{1} \and
Peibei Shi\inst{2} \and
Christoph Bergmeir\inst{1} \and
Hansika Hewamalage \inst{1} \and
Quoc Tran \inst{2} \and
Brian Seaman \inst{2}
}
%

%
\institute{Faculty of Information Technology, Monash University, Melbourne, Australia.
\email{\{herath.bandara,christoph.bergmeir,hansika.hewamalage\}@monash.edu}
\and
@Walmart Labs, San Bruno, USA. \\
\email{\{pshi,qtran,brian\}@walmartlabs.com}
}
\maketitle              
\begin{abstract}
Generating accurate and reliable sales forecasts is crucial in the E-commerce business. The current state-of-the-art techniques are typically univariate methods, which produce forecasts considering only the historical sales data of a single product.  However, in a situation where large quantities of related time series are available, conditioning the forecast of an individual time series on past behaviour of similar, related time series can be beneficial. Since the product assortment hierarchy in an E-commerce platform contains large numbers of related products, in which the sales demand patterns can be correlated, our attempt is to incorporate this cross-series information in a unified model. We achieve this by globally training a Long Short-Term Memory network (LSTM) that exploits the non-linear demand relationships available in an E-commerce product assortment hierarchy. Aside from the forecasting framework, we also propose a systematic pre-processing framework to overcome the challenges in the E-commerce business. We also introduce several product grouping strategies to supplement the LSTM learning schemes, in situations where sales patterns in a product portfolio are disparate. We empirically evaluate the proposed forecasting framework on a real-world online marketplace dataset from Walmart.com. Our method achieves competitive results on category level and super-departmental level datasets, outperforming state-of-the-art techniques.

\end{abstract}
\section{Introduction}
\label{sec:intro}
Generating product-level demand forecasts is a crucial factor in E-commerce platforms. Accurate and reliable demand forecasts enable better inventory planning, competitive pricing, timely promotion planning, etc. While accurate forecasts can lead to huge savings and cost reductions, poor demand estimations are proven to be costly in this domain. 

\begin{figure}[htbp]
\centerline{\includegraphics[width=0.56\textwidth]{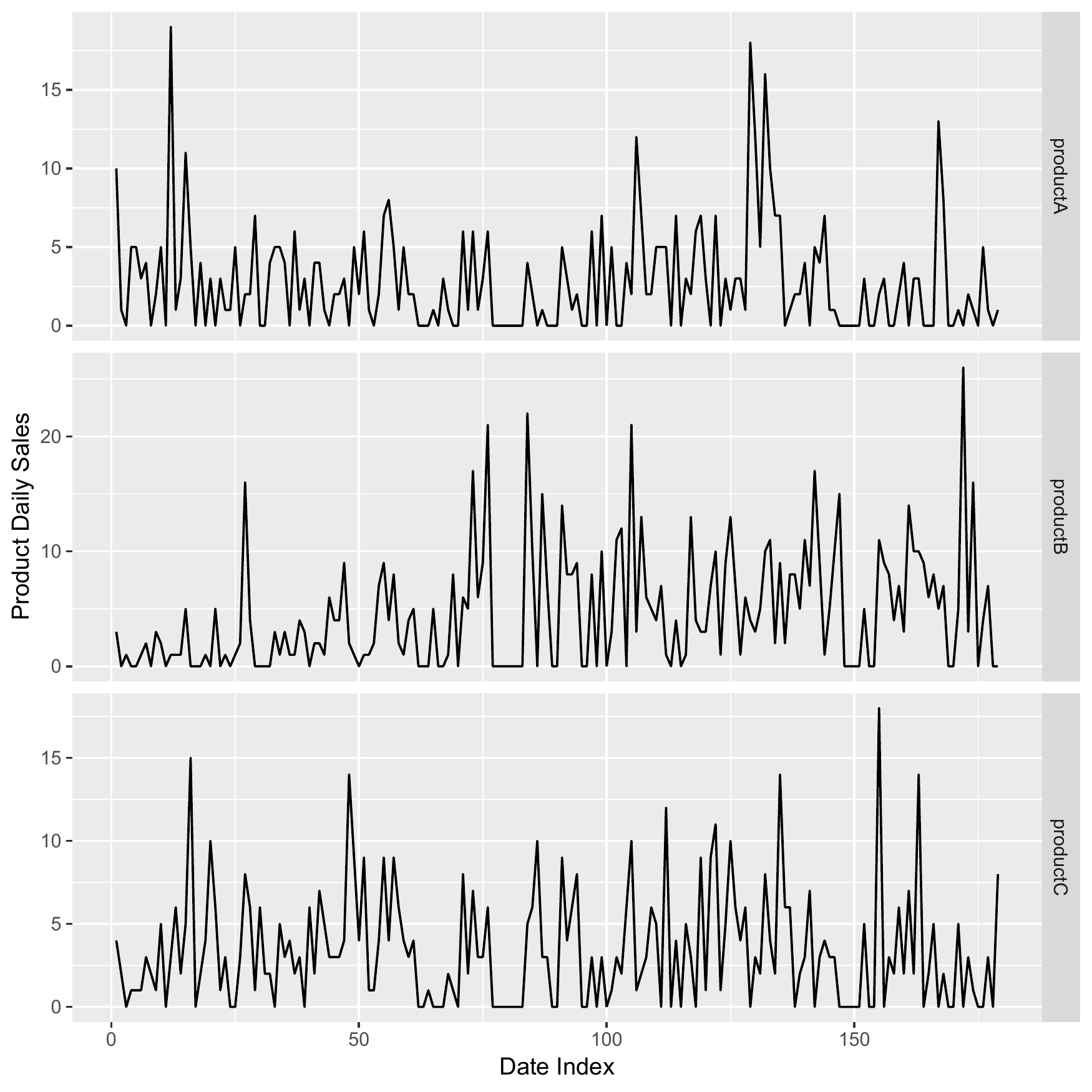}}
\caption{Daily sales demand of three different products over a four months period, extracted from \textit{Walmart.com}. These products are collected from the same product assortment sub-hierarchy.}\label{timeseries}
\end{figure}

The business environment in E-commerce is highly dynamic and often volatile, which is largely caused by holiday effects, low product-sales conversion rate, competitor behaviour, etc. As a result, demand data in this domain carry various challenges, such as highly non-stationary historical data, irregular sales patterns, sparse sales data, highly intermittent sales, etc. Furthermore, product assortments in these platforms follow a hierarchical structure, where certain products within a subgroup of the hierarchy can be similar or related to each other. In essence, this hierarchical structure provides a natural grouping of the product portfolio, where items that fall in the same subcategory/category/department/su
per-department are considered as a single group, in which the sales patterns can be correlated. The time series of such related products are correlated and may share key properties of demand. For example, increasing demand of an item may potentially cause to decrease/increase sales demand of another item, i.e., substituting/complimentary products. Therefore, accounting for the notion of similarity between these products becomes necessary to produce accurate and meaningful forecasts in the E-commerce domain. An example of such related time series is shown in Fig.~\ref{timeseries}.

The existing demand forecasting methods in the E-commerce domain are largely influenced by state-of-the-art forecasting techniques from the exponential smoothing \cite{b1} and the ARIMA \cite{b2} families. However, these forecasting methods are univariate, thus treat each time series separately, and forecast them in isolation. As a result, though many related products are available, in which the sales demand patterns can be similar, these univariate models ignore such potential cross-series information available within related products. Consequently, efforts to exploit the enormous potentials of such multiple related time series is becoming increasingly popular \cite{b13,b14,b15,b16,b17,b18}. More recently, Recurrent Neural Networks (RNN) and Long Short-Term Memory Networks (LSTM), a special group of neural networks (NN) that are naturally suited for time series forecasting, have achieved promising results by globally training the network across all related time series that enables the network to exploit any cross-series information available \cite{b15,b16,b18}.

Therefore, with the primary objective of leveraging demand forecasts in the E-commerce domain, we identify the main research contributions of this study as follows:

\begin{itemize}
\item We exploit sales correlations and relationships available in an E-commerce product hierarchy. 
\item We introduce a systematic preprocessing framework to overcome the data challenges in the E-commerce domain.
\item We analyse and compare two different LSTM learning schemes with different back-propagation error terms, and include a mix of static and dynamic features to incorporate potential external driving factors of sales demand.
\item We empirically evaluate our framework using real-world retail sales data from \textit{Walmart.com}, in which we use state-of-the-art forecasting techniques to compare against our proposed framework.
\end{itemize}

The rest of the paper is organized as follows. In Section~\ref{sec:problem} we formally define the problem of generating a global time series model for product demand forecasting. In Section~\ref{sec:relatedwork} we discuss the state of the art in this domain. We describe the proposed preprocessing scheme in Section~\ref{sec:preprocess}. Next, in Section~\ref{sec:network}, we outline the key learning properties included in our LSTM network architecture. We summarise the overall architecture of our forecasting engine in Section~\ref{sec:overall}. Our experimental setup is presented in Section~\ref{sec:experiments}, where we demonstrate the results obtained by applying our framework to a large dataset from \textit{Walmart.com}. Finally, Section~\ref{sec:conclusion} concludes the paper.

\section{Problem Statement}
\label{sec:problem}
Let $i\ \in \{1, 2, ..., n\}$ be the $i$th product from $n$ total products in our database. The historical sales of product $i$ are given by $X_i = \{x_1, x_2, ..., x_K\} \in {\Bbb R^{K}}$, where $K$ represents the length of a time series. Additionally, we introduce an exogenous feature space, $Z_i = \{z_1, z_2, ..., z_K\} \in {\Bbb R^{K \times P}}$, where $P$ denotes the feature dimension of $Z_i$.

Our aim is to develop a prediction model $f$, which uses the past sales data of all the products in the database, i.e., $X = \{X_1, X_2, ..., X_n\} \in {\Bbb R^{n \times K}}$, and the exogenous feature set $Z = \{Z_1, Z_2, ..., Z_n\} \in {\Bbb R^{Kn \times P}}$ to forecast $M$ number of future sales demand points of product $i$, i.e., $X^M_{i} = \{x_t, x_{t+1}, ..., x_{t+M}$\}, where $M$ is the forecasting horizon. The model $f$ can be defined as follows:
\begin{equation}
  X^M_{i} = f(X,Z,\theta)
  \label{eq0}
\end{equation}

Here, $\theta$ are the model parameters, which are learned in the LSTM training process.

\section{Prior Work}
\label{sec:relatedwork}
The traditional demand forecast algorithms are largely influenced by state-of-the-art univariate statistical forecasting methods such as exponential smoothing methods \cite{b1} and ARIMA models \cite{b2}. As described earlier, forecasting in the E-commerce domain commonly needs to address challenges such as irregular sales patterns, presence of highly bursty and sparse sales data, etc. Nonetheless, numerous studies have been
undertaken to alleviate the limitations of classical approaches in these challenging conditions. This includes introducing preprocessing techniques \cite{b3}, feature engineering methods \cite{b4,b5,b6,b7}, and modified likelihood functions \cite{b8,b9}.

As emphasized in Section~\ref{sec:intro}, one major limitation of univariate forecasting techniques is that they are incapable of using cross-series information for forecasting. Also many studies based on NNs, which are recognised as a strong alternative to traditional approaches, have been employing NNs in the form of a univariate forecasting technique \cite{b10,b11,b12}. 

In addition to improving the forecasting accuracy, forecasting models that build on multiple related time series can be potentially more robust in handling outliers in a time series. This is because, incorporating the common behaviour of multiple time series may reduce the effects caused by few abnormal observations in a single time series.

Recently, methods to build global models across such time series databases have achieved promising results. Trapero et al. \cite{b13} introduce a pooling regression model on sets of related time series. They improve the promotional forecast accuracy in situations where historical sales data is limited in a single time series. Chapados \cite{b17} achieves good results in the supply chain planning domain by modelling multiple time series using a Bayesian framework, where that author uses the available hierarchical structure to disseminate the cross-series information across a set of time series. More recently, deep learning techniques, such as RNNs and CNNs have also shown to be competitive in this domain \cite{b14,b15,b16,b18}.

The probabilistic forecasting framework introduced by \cite{b15,b16} attempts to address the uncertainty factor in forecasting. Those authors use RNN and LSTM architectures to learn from groups of time series, and provide quantile estimations of the forecast distributions. Moreover, Bandara et al. \cite{b18} develop a clustering-based forecasting framework to accommodate situations where groups of heterogeneous time series are available. Here, those authors initially group the time series into subgroups based on a similarity measure, before using RNNs to learn across each subgroup of time series. Furthermore, Borovykh et al. \cite{b14} apply CNNs to model similar sets of financial time series together, where they highlight that the global learning procedure improves both robustness and forecasting accuracy of a model, and also enables the network to effectively learn from shorter time series, where information available within an individual time series is limited.

\section{Data Preprocessing}
\label{sec:preprocess}
Sales datasets in the E-commerce domain experience various issues that we aim to address with the following preprocessing mechanisms in our framework.

\subsection{Handling Data Quality Issues}
\label{sec: fixdqi}

Nowadays, data extract, transform, load (ETL) \cite{b37} is the main data integration process in data warehousing pipelines. However, the ETL process is often unstable in real-time processing, and may cause false ``zero" sales in the dataset. Therefore, we propose a method to distinguish the actual zero sales from the false zero sales (``fake zeros'') and treat the latter as missing observations.

Our approach is mostly heuristic, where we initially compute the minimum non-zero sales of each item in the past 6 months. Then, we treat the zero sales as ``fake'' zero sales if the minimum non-zero sales of a certain item are higher than a threshold $\gamma$. We treat these zero sales as missing observations. It is also noteworthy to mention that the ground truth of zero sales is not available, thus potential false positives can appear in the dataset.

\subsection{Handling Missing Values}
\label{sec: missval}
We use a forward-filling strategy to impute missing sales observations in the dataset. This approach uses the most recent valid observation available to replace the missing values. We performed preliminary experiments that showed that this approach outperforms more sophisticated imputation techniques such as linear regression and Classification And Regression Trees (CART).

\subsection{Product Grouping}
\label{sec:itemgroup}
According to \cite{b18}, employing a time series grouping strategy can improve the LSTM performance in situations where time series are disparate. Therefore, we introduce two product grouping mechanisms in our preprocessing scheme.

In the first approach, the target products are grouped based on available domain knowledge. Here, we use the sales ranking and the percentage of zero sales as primary business metrics to form groups of products. The first group (G1) represents the product group with a high sales ranking and a low zero sales density. Whereas, group 2 (G2) represents the product group with a low sales ranking and a high zero sales density. Group 3 (G3) represents the rest of the products. From an E-commerce perspective, products in G1 are the ``head items'' that bring the highest contribution to the business, thus improving the sales forecast accuracy in G1 is most important. Details of the above groupings are summarized in Table \ref{itemgrouping}. 

The second approach is based on time series clustering, where we perform K-means clustering on a set of time series features to identify the product grouping. Table \ref{tsfeature} provides an overview of these features, where the first two features are business specific features, and the rest are time series specific features. The time series features are extracted using the \textit{tsfeatures} package developed by \cite{b36}. Finally, we use a \textit{silhouette analysis} to determine the optimal number of clusters in the K-means clustering algorithm.

\begin{table}
\caption{Time series and sales-related features used for product clustering}
\begin{center}
\begin{tabular}{ll}
\hline
Feature    &Description \\ 
\hline
Sales.quantile  &Sales quantile over total sales \\
Zero.sales.percentage &Sales sparsity/percentage of zero sales\\
Trend &Strength of trend  \\
Spikiness 	&Strength of spikiness  \\
Linearity 	&Strength of linearity \\
Curvature 	&Strength of curvature \\
ACF1-e &Autocorrelation coefficient at lag 1 of the residuals\\
ACF1-x &Autocorrelation coefficient at lag 1\\
Entropy & Spectral entropy \\
\hline
\end{tabular}
\label{tsfeature}
\end{center}
\end{table}
\vspace*{-10mm}
\begin{table} 
\caption{Sales sparsity thresholds used for domain-based product grouping}
\begin{center}
\begin{tabular}{lll}
\hline
Group-ID &Sales ranking &Sales sparsity\\
\hline
1 &Sales.quantile $\leq$ 0.33 &Zero.sales.percentage.quantile $\geq$ 0.67\\
2 &Sales.quantile $\geq$ 0.67 &Zero.sales.percentage.quantile $\leq$ 0.33 \\
3 &other &other \\
\hline
\end{tabular}
\label{itemgrouping}
\end{center}
\vspace{-0.6cm}
\end{table} 

\subsection{Sales Normalization}
\label{sec:datanormal}
The product assortment hierarchy is composed of numerous commodities that follow various sales volume ranges, thus performing a data normalisation strategy becomes necessary before building a global model like ours. We use the \textit{mean-scale} transformation proposed by \cite{b15}, where the mean sales of a product is considered as the scaling factor. This transformation can be formally defined as follows:


\begin{equation}
X_{i, new} = \frac{X_i}{1+ \frac{1}{k}\sum_{t=1}^{k}{X_{i,t}}}
\label{meanscale}
\end{equation}

Here, $X_{i, new}$ represents the normalised sales vector, and $k$ denotes the number of sales observations of product $i$.

\subsection{Moving Window Approach}
\label{sec:mwa}

The Moving Window (MW) strategy transforms a time series ($X_i$) into pairs of $<$\textit{input, output}$>$ patches, which are later used as the training data of the LSTM. 

\begin{figure}[htbp]
\centerline{\includegraphics[width=0.65\textwidth]{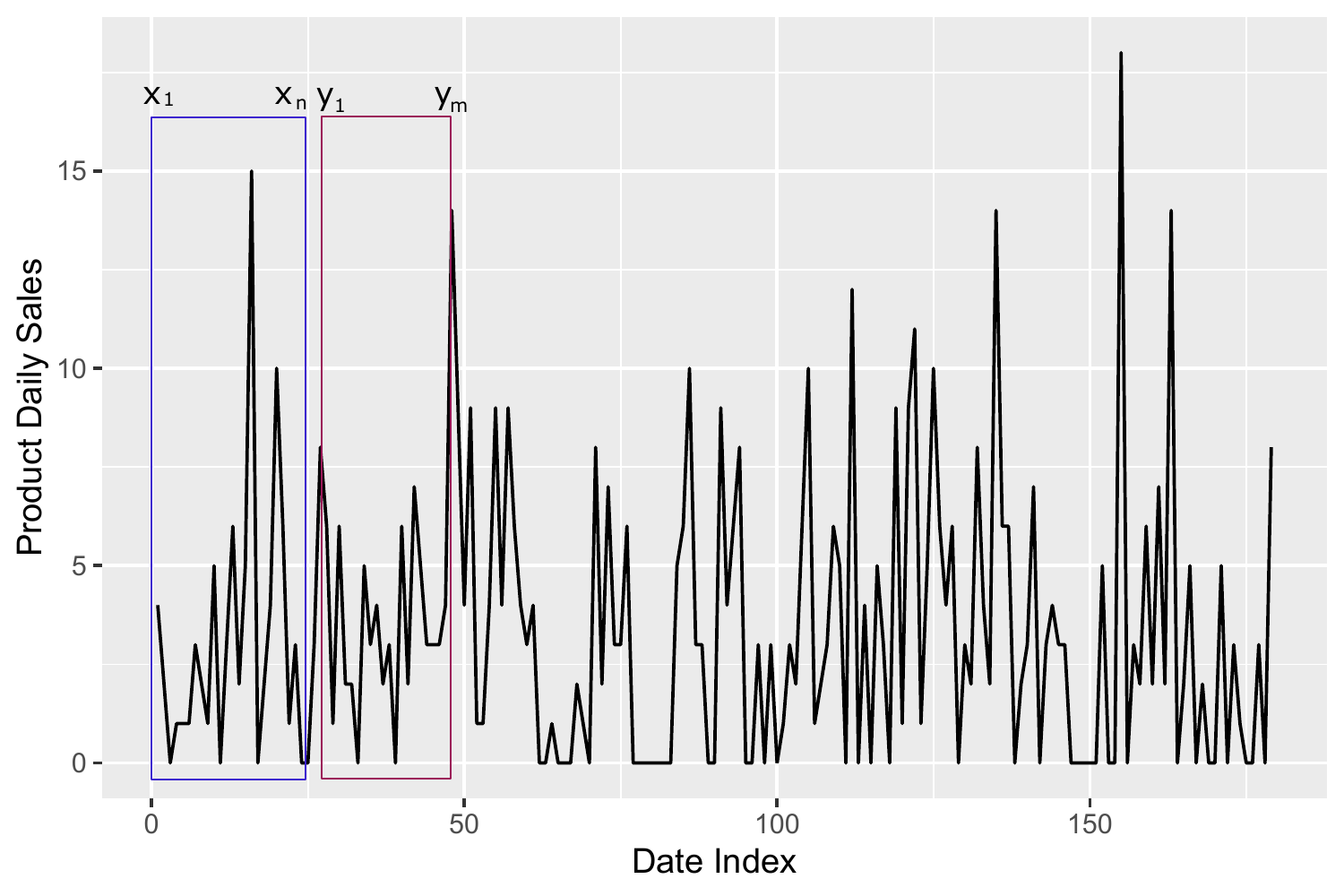}}
\caption{Applying the MW approach to time series $X_i$. Here, $\{x_1, x_2, ..., x_n\}$ refers to the input window, and $\{y_1, y_2, ..., y_m\}$ is the corresponding output window.}
\label{mw}
\end{figure}

Given a time series $X_i$ = \{$x_1, ... ,x_K$\} $\in$ ${\Bbb R^{K}}$ of length $K$, the MW strategy converts the $X_i$ into $(K-n-m)$ number of patches, where each patch has a size of $(m + n)$. Here, $n$ and $m$ represent the sizes of the input window and output window, respectively. In our study, we make the size of the output window ($m$) identical to the intended forecasting horizon, following the Multi-Input Multi-Output (MIMO) strategy in multi-step forecasting. This enables our model to directly predict all future values up to the intended forecasting horizon $X^M_{i}$. The MIMO strategy is advocated by many studies \cite{b16,b27} for multi-step forecasting with NNs.
Fig.~\ref{mw} illustrates an example of applying the MW approach to a sales demand time series from our dataset.

We use an amount of $(K-m)$ data points from time series $X_i$ to train the LSTM, and reserve the last output window of $X_i$ for the network validation. To avoid possible network saturation effects, which are caused by the bounds of the network activation functions \cite{b28}, we employ a local normalisation process at each MW step. In this step, the mean value for each input window ($\bar{X_{i}}$) is calculated and subtracted from each data point of the corresponding input and output window. Thereafter, these windows are shifted forward by one step, i.e., $\{x_2, x_3, ..., x_{n+1}\}$, $\{y_2, y_3, ..., y_{m+1}\}$, and the normalisation process is repeated. The normalisation procedure also enables the network to generate conservative forecasts (for details see Bandara et al.~\cite{b18}), which is beneficial in forecasting in general, and in particular in the E-commerce domain, as this reduces the risk of generating large demand forecasting errors.

\section{LSTM Network Architecture}
\label{sec:network}

LSTMs are an extension of RNNs that have the ability to learn long-term dependencies in a sequence, overcoming the limitations of vanilla RNNs \cite{b21}. The cohesive gating mechanism, i.e., input, output, and forget gates, together with the self-contained memory cell, i.e., ``Constant Error Carousel'' (CEC) allow the LSTM to regulate the information flow across the network. This enables the LSTM to propagate the network error for much longer sequences, while capturing their long-term temporal dependencies.

In this study, we use a special variant of LSTMs, known as ``LSTM with peephole connections'' that requires the LSTM input and forget gates to incorporate the previous state of the LSTM memory cell. For further discussions of RNN and LSTM architectures, we refer to \cite{b18}.
In the following, we describe how exactly the LSTM architecture is used in our work.

\subsection{Learning Schemes}
As mentioned in Section \ref{sec:mwa}, we use the input and output data frames generated from the MW procedure as the primary training source of LSTM. Therefore, the LSTM is provided with an array of lagged values as the input data, instead of feeding in a single observation at a time. This essentially relaxes the LSTM memory regulation and allows the network to learn directly from a lagged time series \cite{b18}. 

\begin{figure}[htp]
\subfloat[An unrolled representation of learning scheme LSTM-LS1]{%
  \centerline{\includegraphics[width=0.50\textwidth]{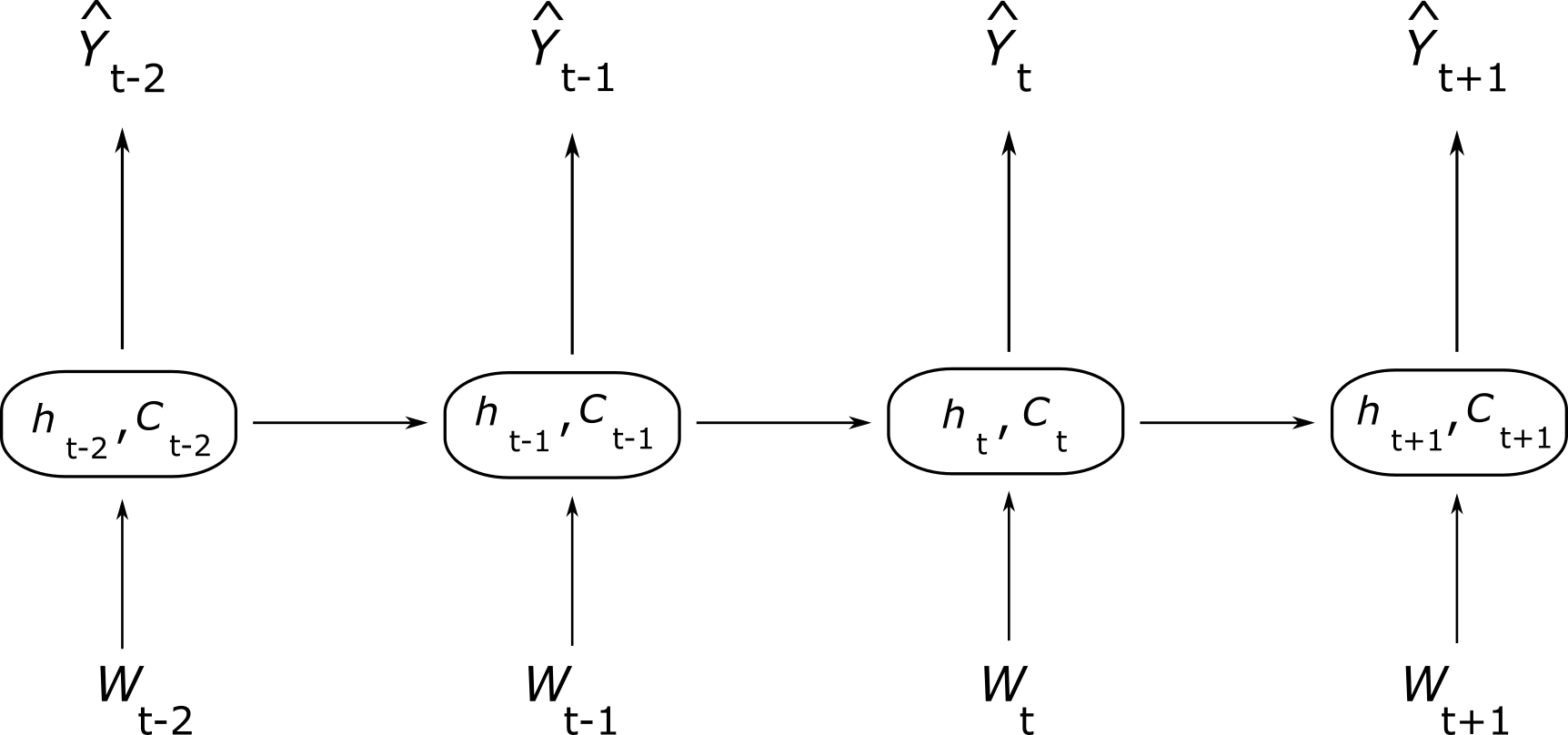}}
}
\vspace{4mm}
\subfloat[An unrolled representation of learning scheme LSTM-LS2]{%
  \centerline{\includegraphics[width=0.50\textwidth]{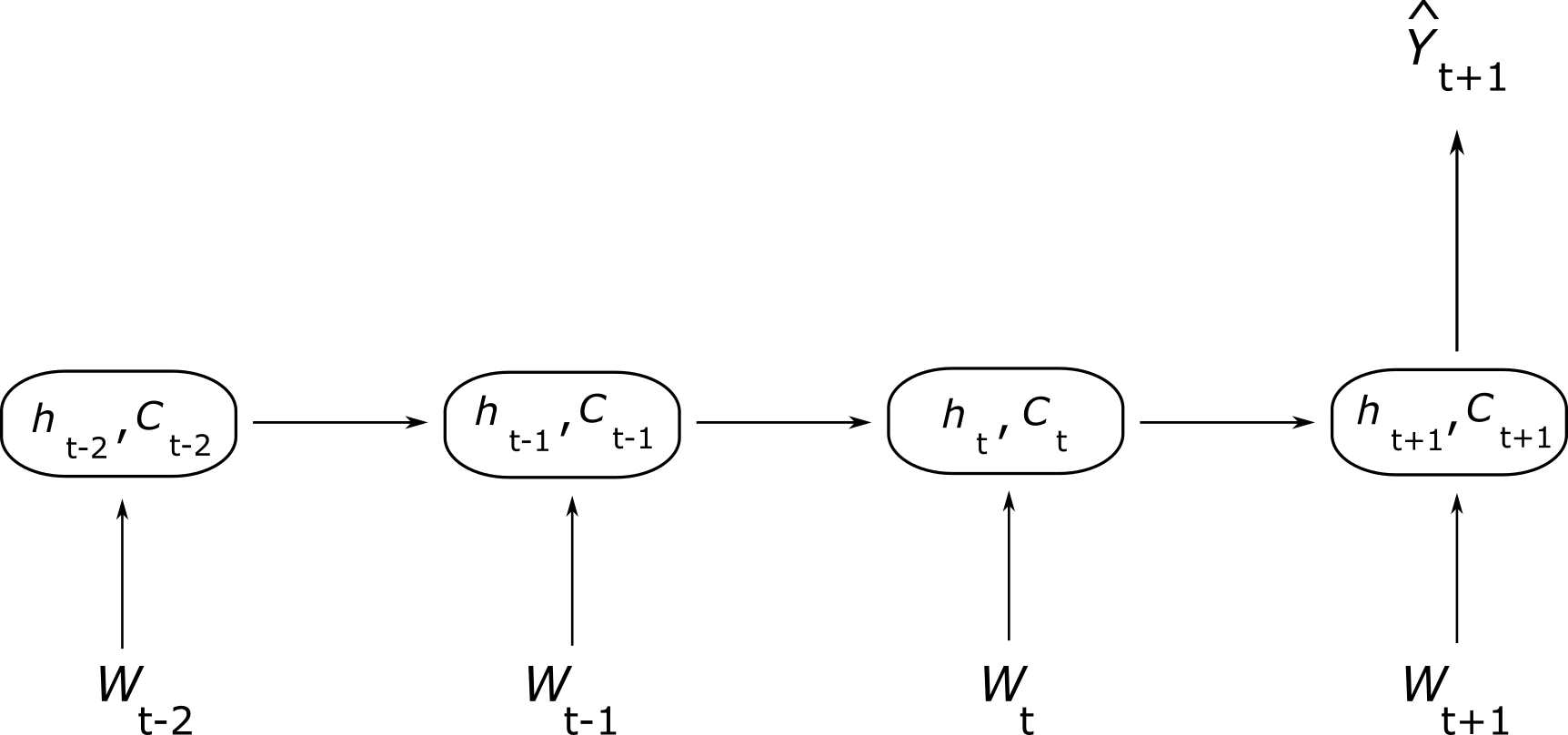}}
}
\caption{The architectures of LSTM learning schemes, LSTM-LS1 and LSTM-LS2. Each squared unit represents a peephole connected LSTM cell, where $h_t$ provides short-term memory and $C_t$ retains the long-term dependencies of LSTM.}
\label{lstm-schemes}
\end{figure}

Fig.~\ref{lstm-schemes} summarizes the LSTM learning schemes used in our study, LSTM-LS1 and LSTM-LS2. Here,  $W_{t}$ $\in$ ${\Bbb R^{n}}$ represents the input window at time step $t$, $h_t$ $\in$ ${\Bbb R^{p}}$ represents the hidden state at time step $t$, and the cell state at time step $t$ is represented by $C_{t-1}$ $\in$ ${\Bbb R^{p}}$. Note that $p$ denotes the dimension of the memory cell of the LSTM. Additionally, we introduce $\hat{Y_{t}}$ $\in$ ${\Bbb R^{m}}$ to represent the projected output of the LSTM at time step $t$. Here, $m$ denotes our output window size, which is equivalent to the forecasting horizon $M$. Here, each LSTM layer is followed by a fully connected neural layer (excluding the bias component) to project each LSTM cell output $h_t$ to the dimension of the output window $m$.

The proposed learning schemes can be distinguished by the overall error term $E_t$ used in the network back-propagation, which is back-propagation through time (BPTT;\cite{b20}). Given ${Y_{t}}$ $\in$ ${\Bbb R^{m}}$ are the actual observations of values in the output window at time step $t$, which are used as the teacher inputs for the predictions $\hat{Y_{t}}$, the LSTM-LS1 scheme accumulates the error $e_t$ of each LSTM cell instance to compute the error $E_t$ of the network. Here, $e_t$ refers to the prediction error at time step $t$, where $e_t=Y_t-\hat{Y_{t}}$. Whereas in LSTM-LS2, only the error term of the final LSTM cell instance $e_{t+1}$ is used as the error $E_t$ for the network training. For example, in Fig.~\ref{lstm-schemes}, the $E_t$ of LSTM-LS1 scheme is equivalent to $\sum_{j=t-2}^{t+1}e_{j}$, while the error term in the final LSTM cell state $e_{t+1}$ gives the error $E_t$ of LSTM-LS2. These error terms are eventually used to update the network parameters, i.e., the LSTM weight matrices. In this study, we use TensorFlow, an open-source deep-learning toolkit \cite{b29} to implement the above LSTM learning schemes.

\begin{figure}[htbp]
\centerline{\includegraphics[width=0.65\textwidth]{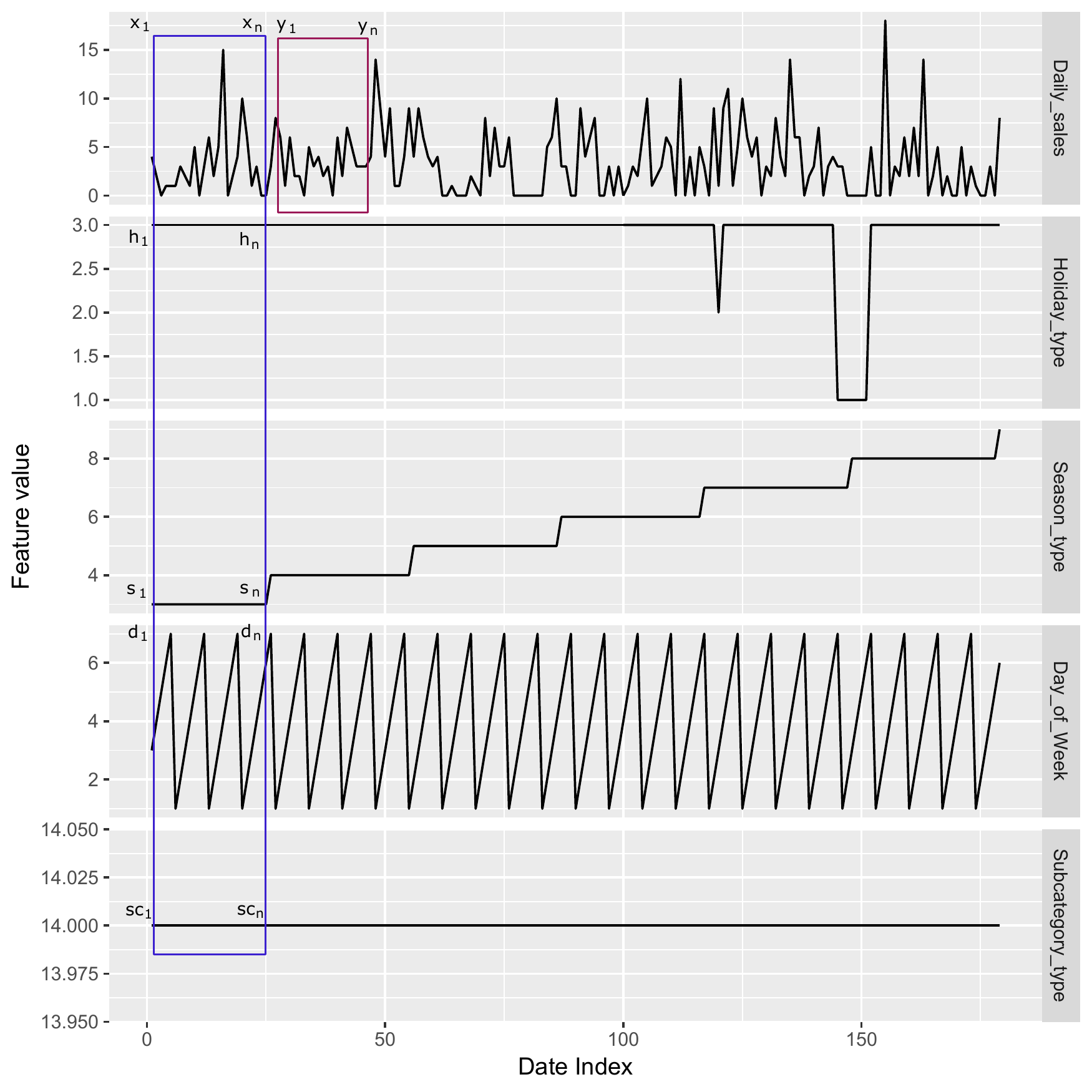}}
\caption{Using both static $Z_t^{(s)}$, and dynamic $Z_t^{(d)}$ features with the MW approach. All categorical variables are represented as ``one-hot-encoded'' vectors in the LSTM training data.}
\label{featuremw}
\end{figure}

\vspace*{-10mm}

\subsection{Exogenous Variables}
We use a combination of static and dynamic features to model external factors that affect the sales demand. In general, static features include time invariant information, such as product type, product category, etc. Dynamic features include calendar features (e.g., holidays, season, weekday/weekend). These features can be useful to capture sales demand behaviours of products in a certain period of time.

Fig.~\ref{featuremw} demonstrates an example of applying the MW approach (see Section~\ref{sec:mwa}) to include static and dynamic features in an input window. Now, the input window $W_{t}$ is a unified vector of past sales observations $X_t$, static features $Z_t^{(s)}$, and dynamic features $Z_t^{(d)}$. As a result, in addition to historical sales observations $\{x_1, x_2, ..., x_n\}$, we also include the input windows of the holidays $\{h_1, h_2, ..., h_n\}$, seasons $\{s_1, s_2, ..., s_n\}$, day of the week $\{d_1, d_2, ..., d_n\}$, and the sub category types $\{sc_1, sc_2, ..., sc_n\}$. Later, LSTM uses a concatenation of these input windows to learn the actual observation of the output window $\{y_1, y_2, ..., y_m\}$.

\section{Overall procedure}
\label{sec:overall}

\begin{figure*}[htbp]
\centerline{\includegraphics[width=1.0\textwidth]{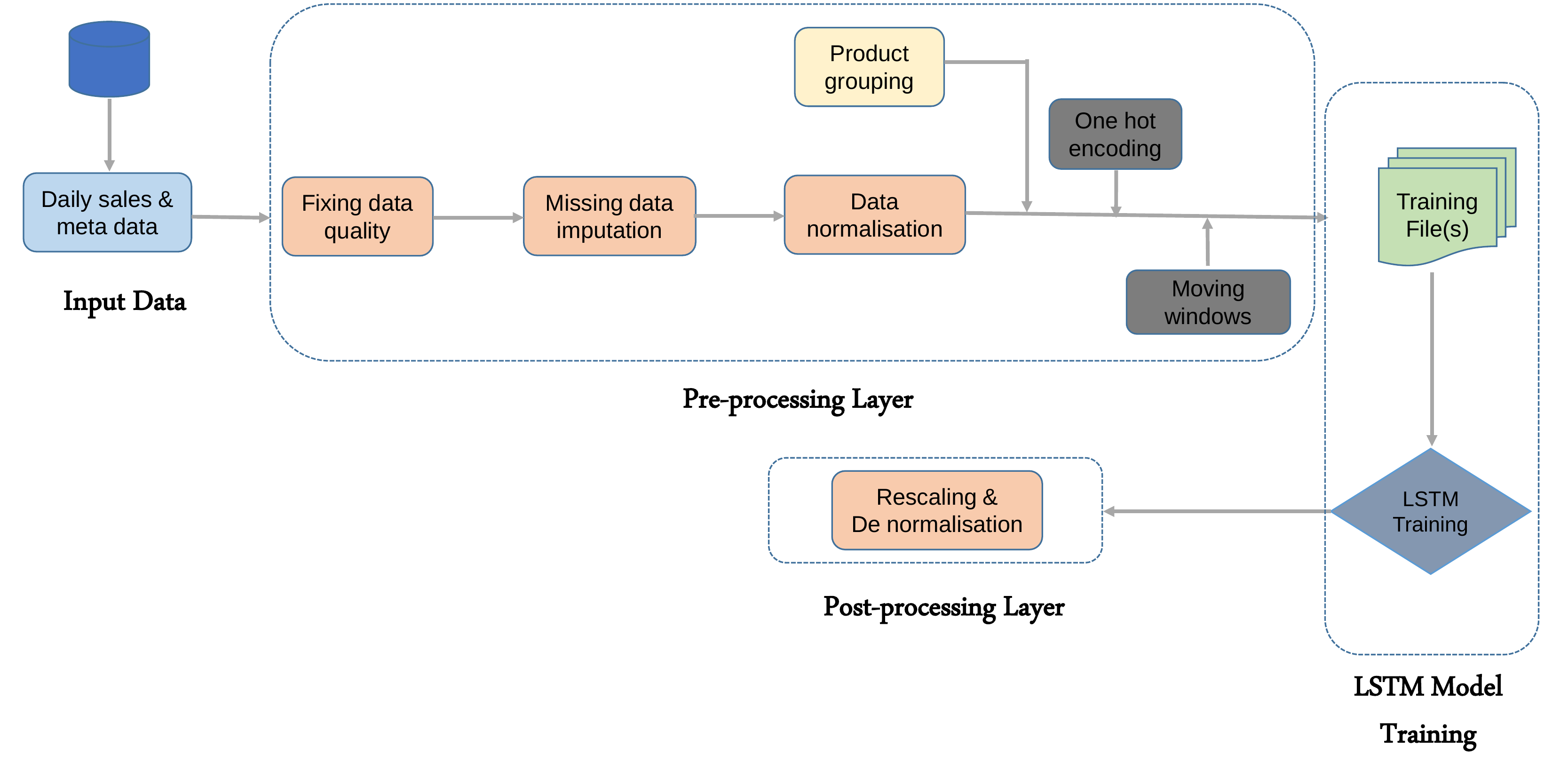}}
\caption{The overall summary of the proposed sales demand forecasting framework, which consists of a pre-processing, an LSTM training, and a post-processing part.}
\label{process}
\end{figure*}

The proposed forecasting framework is composed of three components, namely 1) pre-processing layer, 2) LSTM training layer, and 3) post-processing layer. Fig.~\ref{process} gives a schematic overview of our proposed forecasting framework. 

As described in Section~\ref{sec:preprocess}, we initially perform several preprocessing techniques to arrange the raw data for the LSTM training procedure. Afterwards, the LSTM models are trained according to the LSTM-LS1 and LSTM-LS2 learning schemes shown in Fig.~\ref{lstm-schemes}. Then, in order to obtain the final forecasts, we rescale and denormalize the predictions generated by the LSTM. Here, the rescaling process back-transforms the generated forecasts to their original scale of sales, whereas the denormalization process (see Section~\ref{sec:mwa}) adds back the mean sales of the last input window to the forecasts.

\vspace{-5mm}
\section{Experiments}
\label{sec:experiments}
In this section, we describe the experimental setup used to empirically evaluate our proposed forecasting framework. This includes the datasets, error metrics, hyper-parameter selection method, benchmark methods and LSTM variants used to perform the experiments, and the results obtained.

\subsection{Datasets}
We evaluate our forecasting framework on two datasets collected from \textit{Walmart.com}. We first evaluate our framework on a subset of 1724 items that belong to the product \textit{household} category, which consists of 15 different sub-categories. Next, we scale up the number of products to 18254 by extracting a collection from a single super-department, which consists of 16 different categories.

We use 190 consecutive days of daily sales data in 2018. The last 10 days of data are reserved for model testing. We define our forecasting horizon $M$ as 10, i.e., training output window size $n$ is equivalent to 10. Following the heuristic proposed by Bandara et al. \cite{b18}, we choose the size of the training input window $n$ as 13 (10*1.25).

\begin{table}
\caption{LSTM Parameter grid}
\begin{center}
\begin{tabular}{lcc}
	\toprule
	Model Parameter            			&Minimum value 			&Maximum value\\ \hline
	LSTM-cell-dimension  				&50						&100	\\
	Mini-batch-size						&60						&1500	\\
	Learning-rates-per-sample			&$10^{-6}$				&$10^{-3}$	\\
	Maximum-epochs						&5						&20	\\
	Gaussian-noise-injection			&$10^{-4}$				&$8 \cdot 10^{-4}$
\\
	L2-regularization-weight			&$10^{-4}$				&$8 \cdot 10^{-4}$\\ \hline
\end{tabular}
\label{tab:parametergrid}
\end{center}
\vspace{-0.3cm}
\end{table}

\subsection{Error Measure}
We use a modified version of the mean absolute percentage error (mMAPE) as our forecasting error metric. We define the mMAPE for each item as:

\begin{equation}
\text{mMAPE} = \frac{1}{m}\sum_{t=1}^{m}\left(\frac{\left|F_t - A_t\right|}{1 + \left| A_t\right|} \right).
\label{mape}
\end{equation}

Here, $A_t$ represents the actual sales at time $t$, and $F_t$ is the respective sales forecast generated by a prediction model. The number $m$ denotes the length of the intended forecasting horizon. Furthermore, to avoid problems for zero values, we sum a constant term $\epsilon=1$ to the denominator of \eqref{mape}. This is a popular error measure used in the retail and supply chain industry \cite{b39}.

To report the overall mMAPE for a set of items, we use both mean of the mMAPEs (Mean mMAPE) and the median of the mMAPEs (Median mMAPE).
Here, median mMAPE is suitable to summarise the error distribution in situations where the majority of the observations are zero sales, i.e., long tailed sales demand items.

\subsection{Hyperparameter Selection \& Optimization}
Our LSTM based learning framework contains various hyper-parameters, including LSTM cell dimension, model learning rate, number of epochs, mini-batch-size, and model regularization terms, i.e., Gaussian-noise and L2-regularization weights. We use two implementations of a Bayesian global optimization methodology, \textit{bayesian-optimization} and \textit{SMAC} \cite{b30} to autonomously determine the optimal set of hyper-parameters in our model \cite{b32}. Table \ref{tab:parametergrid} summarises the bounds of the hyper-parameter values used throughout the LSTM learning process, represented by the respective minimum and maximum columns.

Moreover, we use the gradient-based \textit{Adam} \cite{b33} and \textit{COntinuous COin Betting (COCOB)} \cite{b34} algorithms as our primary learning optimization algorithms to train the network. Unlike in other gradient-based optimization algorithms, \textit{COCOB} does not require tuning of the learning rate.

\subsection{Benchmarks and LSTM variants}
We use a host of different univariate forecasting techniques to benchmark against our proposed forecasting framework. This includes forecasting methods from the exponential smoothing family, i.e., \textit{exponentially weighted moving average (EWMA)}, \textit{exponential smoothing (ETS)} \cite{b35}, and a model from the moving average family, i.e., \textit{autoregressive moving-average model (ARIMA) \cite{b35}}. Though some of these benchmarks have been proposed in the literature decades ago, they are used in many businesses as the forecasting work-horses on a daily basis, and recent forecasting competitions have shown that even today these methods are able to obtain very competitive accuracies \cite{b38}. We also use \textit{Prophet}, a forecasting technique recently introduced by Facebook Research \cite{b40}, as a benchmark. In addition to the well-established benchmarks in this domain, we include standard benchmarks such as Na\"ive, and Na\"ive Seasonal. Some of these benchmarks are also currently used in the forecasting framework at \textit{Walmart.com}. 

Furthermore, in our experiments, we add the following variants of our baseline LSTM model.

\begin{itemize}
\item \textit{LSTM.ALL}: The baseline LSTM model, where one model is globally trained across all the available time series.
\item \textit{LSTM.GROUP}: A separate LSTM model is built on each subgroup of time series, which are identified by the domain knowledge available.
\item \textit{LSTM.FEATURE}: The subgroup labels identified in the \textit{LSTM.GROUP} approach is used as an external feature (one-hot encoded vector) of LSTM.
\item \textit{LSTM.CLUSTER}: The time series sub-grouping is performed using a time series feature based clustering approach (refer Section~\ref{sec:preprocess}). Similar to \\\textit{LSTM.GROUP}, a separate LSTM model is trained on each cluster.
\end{itemize}

\subsection{Results \& Discussion}
Table \ref{householdresult} and Table \ref{superdeptresult} show the results for the category level and super-department level datasets. Here, $k$ corresponds to the number of items in each group, and $G1$/$G2$/$G3$ represent the product sub-groups introduced in Section \ref{sec:itemgroup}. We use a weekly seasonality in the seasonal benchmarks, i.e., ETS (seasonal), Na\"ive Seasonal. It is also worth to mention that for the super-department dataset, we only employ one grouping strategy, namely LSTM.GROUP, and include only the best-performing learning scheme in the category level dataset, which is LSTM-LS1, to examine the robustness of our forecasting framework. 

In the tables, under each LSTM variant, we present the results of the different learning schemes, i.e., LSTM-LS1 and LSTM-LS2, hyper-parameter selection methods, i.e., Bayesian and SMAC, and optimization learning algorithms, i.e., Adam and COCOB, and achieve comparable results. According to Table \ref{householdresult}, considering all the items in the category, the proposed LSTM.Cluster variant obtains the best Mean mMAPE, while the Na\"ive forecast gives the best Median mMAPE. Meanwhile, regarding G1, which are the items with most business impact, the LSTM.Cluster and LSTM.Group variants outperform the rest of the benchmarks, in terms of the Mean mMAPE and Median mMAPE respectively. We also observe in G1 that the results of the LSTM.ALL variant are improved after applying our grouping strategies. Furthermore, on average, the LSTM variants together with the Na\"ive forecast achieve the best-performing results within G2 and G3, where the product sales are relatively sparse compared to G1. 

We observe a similar pattern of results in Table \ref{superdeptresult}, where in general, the LSTM.GROUP variant gives the best Mean mMAPE, while the Na\"ive forecast ranks as the first in Median mMAPE. Likewise in G1, the LSTM.GROUP variant performs superior amongst other benchmarks, and in particular outperforms the LSTM.ALL variant, while upholding the benefits of item grouping strategies under these circumstances. Similarly, on average, the LSTM variants and Na\"ive forecast obtain the best results in G2 and G3. In both tables, we observe several methods producing zero Median mMAPE in the G2 subgroup. This is due to the high volume of zero sales present among the items in G2. In E-commerce business, items in the G2 are called ``tail items", which are usually seasonal products. These items follow low sales during most time of a year and high sales during certain period of time. Therefore, generating demand forecast for these items is still essential, although their sales are sparse.

Overall, the majority of the LSTM variants show competitive results under both evaluation settings, showing the robustness of our forecasting framework with large amounts of items. More importantly, these results reflect the contribution made by the time series grouping strategies to uplift the baseline LSTM performance.

\vspace*{-4mm}

\begin{table*}[!htb]
\caption{Results for category level dataset}
\begin{center}
\resizebox{\columnwidth}{!}{
\begin{tabular}{llcccccccc}
\toprule
 & & \multicolumn{2}{c}{\scriptsize{mMAPE (All)}} & \multicolumn{2}{c}{ \scriptsize{mMAPE (G1)}} & \multicolumn{2}{c}{\scriptsize{mMAPE (G2)}}& \multicolumn{2}{c}{\scriptsize{mMAPE (G3)}} \\
  & & \multicolumn{2}{c}{k = 1724} & \multicolumn{2}{c}{k = 549} & \multicolumn{2}{c}{k = 544}& \multicolumn{2}{c}{k = 631} \\
Model & Configuration & Mean & Median & Mean & Median &  Mean & Median & Mean & Median \\
\hline
LSTM.ALL & LSTM-LS1/Bayesian/Adam& 0.888 & 0.328 & 1.872  & 0.692 & 0.110 & 0.073 & 0.640 & 0.283\\
LSTM.ALL & LSTM-LS1/Bayesian/COCOB&  {0.803} & {{0.267}} & 1.762 & 0.791 & {{0.070}} & 0.002 & 0.537 & 0.259 \\
LSTM.ALL & LSTM-LS2/Bayesian/Adam& 0.847 & 0.327 & 1.819 & 0.738 & 0.103 & 0.047 & 0.582 & 0.326  \\
LSTM.GROUP& LSTM-LS1/Bayesian/Adam & 0.873 & 0.302  & 1.882 & \textbf{0.667} & 0.093 & 0.016 & 0.604 & 0.283 \\
LSTM.GROUP & LSTM-LS1/Bayesian/COCOB&1.039 & 0.272 & 2.455 & 0.818 & 0.074 & {\textbf{0.000}} & 0.549 & {\textbf{0.250}} \\
LSTM.GROUP &  LSTM-LS2/Bayesian/Adam& 0.812 & 0.317 & 1.818 & 0.738 & 0.091 & 0.022 & 0.587 & 0.314\\
LSTM.FEATURE &LSTM-LS1/Bayesian/Adam & 1.065 & 0.372 & 2.274 & 0.889 & 0.135 & 0.100 & 0.738 & 0.388\\
LSTM.FEATURE &LSTM-LS1/Bayesian/COCOB& {{0.800}} & {{0.267}} & 1.758 & 0.772 & {\textbf{0.069}}& {\textbf{0.000}} & {{0.533} }& {{0.255}}\\
LSTM.FEATURE & LSTM-LS2/Bayesian/Adam & 0.879 & 0.324 & 1.886 & 0.750 & 0.091 & 0.022 & 0.611 & 0.324 \\
LSTM.CLUSTER &  LSTM-LS1/Bayesian/Adam & 0.954 & 0.313 & 2.109 & 0.869 & 0.135 & 0.110 & 0.625 & 0.322\\
LSTM.CLUSTER &  LSTM-LS1/Bayesian/COCOB & {\textbf{0.793}} & 0.308 &  {\textbf{1.695}} & 0.748 & 0.077 & 0.005 & 0.562 & 0.302\\
LSTM.CLUSTER &  LSTM-LS2/Bayesian/Adam & 1.001 & 0.336 & 2.202 & 0.863 & 0.084 & 0.017 & 0.664 & 0.347 \\
EWMA & \_ & 0.968 & 0.342 & 1.983 & 1.026 & 0.107 & 0.021 & 0.762 & 0.412\\
ARIMA & \_ & 1.153 & 0.677 & 2.322 & 0.898 & 0.103 & 0.056 &  0.730 & 0.496\\
ETS (non-seasonal) & \_ & 0.965 & 0.362 & 2.020 & 0.803 & 0.113& 0.060 & 0.713 & 0.444\\
ETS (seasonal) & \_ & 0.983 & 0.363 & 2.070 & 0.804 & 0.116 & 0.059 & 0.713 & 0.445 \\
Na\"ive & \_ & 0.867 & {\textbf{0.250}} & 1.803 & 0.795 & 0.124 & {\textbf{0.000}} & 0.632 & {\textbf{0.250}}\\
Na\"ive Seasonal & \_ &0.811 & 0.347 &  {{1.789}} & {{0.679}} & 0.086 & {\textbf{0.000}} & {\textbf{0.523}} & 0.320 \\
Prophet-Facebook & &  0.892 & 0.342 & 1.923 & 0.842 & 0.103 & 0.042 & 0.609 & 0.325 \\
\hline
\end{tabular}
}
\label{householdresult}
\end{center}
\end{table*}
\vspace*{-10mm}
\begin{table*}[!htb]
\caption{Results for super-department level dataset}
\begin{center}
\resizebox{\columnwidth}{!}{
\begin{tabular}{llcccccccc}
\toprule
& & \multicolumn{2}{c}{ \scriptsize{mMAPE (All items)}} & \multicolumn{2}{c}{\scriptsize{mMAPE (G1)}} & \multicolumn{2}{c}{\scriptsize{mMAPE (G2)}}& \multicolumn{2}{c}{\scriptsize{mMAPE (G3)}} \\
 & & \multicolumn{2}{c}{k = 18254} & \multicolumn{2}{c}{k = 5682} & \multicolumn{2}{c}{k = 5737}& \multicolumn{2}{c}{k = 6835} \\
Model & Configuration & Mean & Median & Mean & Median &  Mean & Median & Mean & Median \\
\hline
LSTM.ALL & LSTM-LS1/Bayesian/Adam  & 1.006 & 0.483 & 2.146 & 1.285 & 0.191 & 0.079 & 0.668 & 0.434\\ 
LSTM.ALL & LSTM-LS1/Bayesian/COCOB & 0.944 & 0.442 & 2.041 & 1.203 & 0.163 & 0.053 & 0.614 & 0.394\\
LSTM.GROUP &LSTM-LS1/Bayesian/Adam &  { \textbf{0.871}} & 0.445 & {\textbf{1.818}} & {\textbf{1.009}} & {0.189} & 0.067 & {\textbf{0.603}} & {{0.377}}\\ 
LSTM.GROUP & LSTM-LS1/Bayesian/COCOB & 0.921 & 0.455 & 1.960 & 1.199 & 0.173 & 0.053 & 0.618 & 0.394\\
LSTM.FEATURE & LSTM-LS1/Bayesian/Adam & {{0.979}} & {{0.424}} & 2.117 & 1.279 & {\textbf{0.151}} & {{0.050}} & {{0.653}} & {{0.377}}\\
LSTM.FEATURE & LSTM-LS1/Bayesian/COCOB &  1.000 & 0.443 & 2.143 & 1.282 & 0.215 & 0.092 & 0.676 & 0.398\\
EWMA & \_&  1.146 & 0.579 & 2.492 & 1.650 & 0.229 & 0.091 & 0.805 & 0.562\\
ARIMA & \_ & 1.084 & 0.536 & 2.305 & 1.497 & 0.198 & 0.094 & 0.734 & 0.510 \\
ETS (non-seasonal) & \_ & 1.097 & 0.527 & 2.314 & 1.494 & 0.204 & 0.092 & 0.755 & 0.509\\
ETS (seasonal) & \_ & 1.089 & 0.528 & 2.290 & 1.483 & 0.204 & 0.092 & 0.756 & 0.510\\
Na\"ive  & \_ & 0.981 & {\textbf{0.363}} & {{2.008}} & {{1.122}} & 0.204  & {\textbf{0.000}} & 0.713 & {\textbf{0.286}}\\
Na\"ive Seasonal & \_ & 1.122 & 0.522 & 2.323 & 1.513& 0.219   & {{0.050}} & 0.803 & 0.475\\
Prophet-Facebook & & 1.087 &  0.554 & 2.266 & 1.400 & 0.210 & 0.113 & 0.765 & 0.534 \\
\hline
\end{tabular}
}
\label{superdeptresult}
\end{center}
\end{table*} 

\vspace*{-10mm}
\section{Conclusions}
\label{sec:conclusion}

There exists great potential to improve sales forecasting accuracy in the E-commerce domain. One good opportunity is to utilize the correlated and similar sales patterns available in a product portfolio. In this paper, we have introduced a novel demand forecasting framework based on LSTMs that exploits non-linear relationships that exist in the E-commerce business.

We have used the proposed approach to forecast the sales demand by training a global model across the items available in a product assortment hierarchy. Our developments also present several systematic grouping strategies to our base model, which are in particular useful in situations where product sales are sparse. 

Our methodology has been evaluated on a real-world E-commerce database from \textit{Walmart.com}. To demonstrate the robustness of our framework, we have evaluated our methods on both category level and super-department level datasets. The results have shown that our methods have outperformed the state-of-the-art univariate forecasting techniques.

Furthermore, the results indicate that E-commerce product hierarchies contain various cross-product demand patterns and correlations are available, and approaches to exploit this information are necessary to improve the sales forecasting accuracy in this domain.

\end{document}